\documentclass[runningheads]{llncs}
\usepackage[T1]{fontenc}
\usepackage{graphicx,verbatim}
\usepackage{multirow}
\usepackage{amssymb}
\usepackage{xcolor}

\begin{document}
%
%\title{Reference- and Memory-Guided Through-Plane Super-Resolution for Cardiac MRI}
\title{Cardiac MRI Through-Plane Super-Resolution Guided by Reference and Memory}
\titlerunning{STRMSR}
%% Removed for anonymized MICCAI submission
\begin{comment} 
\author{First Author\inst{1}\orcidID{0000-1111-2222-3333} \and
Second Author\inst{2,3}\orcidID{1111-2222-3333-4444} \and
Third Author\inst{3}\orcidID{2222--3333-4444-5555}}
%
\authorrunning{F. Author et al.}
\institute{Princeton University, Princeton NJ 08544, USA \and
Springer Heidelberg, Tiergartenstr. 17, 69121 Heidelberg, Germany
\email{lncs@springer.com}\\
\url{http://www.springer.com/gp/computer-science/lncs} \and
ABC Institute, Rupert-Karls-University Heidelberg, Heidelberg, Germany\\
\email{\{abc,lncs\}@uni-heidelberg.de}}
\end{comment} 

\begin{comment} 
\author{Anonymized Authors}  %% Added for anonymized MICCAI 2025 submission
\authorrunning{Anonymized Author et al.}
\institute{Anonymized Affiliations \\
    \email{email@anonymized.com}}
    
\author{Shaoming Pan\inst{1} \and
Chenchuhui Hu\inst{1} \and
Leon Axel\inst{2} \and
Meng Ye\inst{1}\thanks{Corresponding author}}
  %% Added for anonymized MICCAI submission
\authorrunning{Shaoming Pan et al.}
\institute{University of Texas at Arlington \and {New York University Grossman School of Medicine} \\
    \email{sxp0635@mavs.uta.edu}}
\end{comment}

\author{
Shaoming Pan\inst{1} \and
Chenchuhui Hu\inst{1} \and
Leon Axel\inst{2} \and
Meng Ye\inst{1}
}

\authorrunning{Pan et al.}

\institute{
Department of Computer Science and Engineering, University of Texas at Arlington, Arlington, TX, USA \\
\email{\{shaoming.pan, chenchuhui.hu, meng.ye\}@uta.edu}
\and
NYU Grossman School of Medicine, New York University, New York, NY, USA \\
%\email{leon.axel@nyumc.org}
\email{leon.axel@nyulangone.org}
}

\maketitle              % typeset the header of the contribution
\begin{abstract}
Clinical cardiac MRI is commonly acquired with high in-plane resolution but coarse through-plane resolution to reduce scan time and accommodate breath-hold and cardiac-motion constraints, which limits 3D analysis and diagnostic accuracy. We propose \textbf{STRMSR}, a reference- and memory-guided through-plane super-resolution (SR) framework that reconstructs high-resolution (HR) cardiac volumes by leveraging HR reference views acquired from the same subject and intermediate SR results as the memory. 
Our method uses coarse-to-fine contextual matching to establish robust correspondence between low-resolution target and reference/memory images under spatial misalignment. 
A learnable patch-wise dynamic feature aggregation module predicts content-adaptive mixture weights for each local patch, effectively fusing dynamic information while suppressing unreliable feature transfers.
The intermediate SR results stored in the memory bank ensure slice-to-slice consistency for the super-resolved 3D volume.
Experiments on the WHS cardiac MRI dataset under two reference protocols, orthogonal-plane views and long-axis chamber views, demonstrate consistent improvements over baselines at $\times$4 and $\times$8 upsampling factors. Code is available at \url{https://github.com/030108ming/STRMSR}.

% Authors must provide keywords and are not allowed to remove this Keyword section.

\end{abstract}

\section{Introduction}

%Magnetic resonance imaging (MRI) offers non-invasive imaging with detailed anatomical information and superior soft-tissue contrast. However, prolonged 3D acquisitions cause patient discomfort and motion-induced artifacts, which is particularly problematic for cardiac MRI due to breath-hold limits and continuous cardiac motion. Clinical cardiac protocols therefore rely on 2D breath-hold acquisitions with high in-plane resolution but coarse through-plane resolution, and clinically used sequences such as late gadolinium enhancement (LGE) are often acquired with especially coarse through-plane spacing~\cite{dzyubachyk2015super,xia2021super}, limiting downstream 3D analysis and diagnosis~\cite{sui2021gradient}. Standard cardiac MRI protocols also routinely acquire a small set of long-axis views, which provide complementary appearance cues but are challenging to exploit due to cross-view geometric misalignment and sparse coverage of the short-axis stack.

Cardiac MRI (CMR) relies on 2D breath-hold acquisitions with high in-plane resolution but coarse through-plane resolution, limiting downstream 3D analysis and diagnosis~\cite{sui2021gradient}. Standard CMR protocols routinely acquire a small set of long-axis views as complementary appearance cues for the short-axis stack or acquire orthogonal anisotropic 2D image stacks for 3D interpretation from different viewpoints.
Recovering an isotropic 3D volume from such anisotropic stacks is commonly framed as slice-to-volume reconstruction (SVR)~\cite{kuklisova2012reconstruction}, which proceeds in two sequential steps: (i) \emph{slice alignment}, which corrects rigid inter-breath-hold misalignment between 2D image stacks via slice-to-slice or slice-to-volume registration~\cite{villard2016correction,chandler2008correction}, and (ii) \emph{through-plane super-resolution (SR)}, which fuses the aligned anisotropic data into a coherent high-resolution (HR) 3D volume. The registration step has been extensively studied, with mature solutions based on segmentation map intersection matching~\cite{sinclair2017fully,chang2021unsupervised,xu2024improved}. In this work, we focus on the second step and assume the input low-resolution (LR) stacks are already spatially aligned, aiming to learn an accurate through-plane SR model that recovers fine anatomical details using cross-view HR images as guidance.

Early model-based methods formulate SR as an inverse problem regularized by handcrafted priors~\cite{shi2015lrtv,gholipour2010robust,plenge2012super,sui2021gradient}, but rely on linear degradation assumptions and struggle to recover fine details at large upsampling factors. Deep learning has since become the dominant paradigm for MRI SR~\cite{khateri2025mri}, with supervised methods learning LR-to-HR mapping via densely connected CNNs~\cite{chen2018brain}, multiscale convolutional networks~\cite{pham2019multiscale,choi2025tesla}, and transformers~\cite{li2022transformer}. However, most existing approaches operate on single-input or implicitly fuse multi-input data, limiting their ability to exploit complementary structural information from auxiliary views, and typically process slices independently, leading to inter-slice discontinuities in the reconstructed volume. Self-supervised methods such as SMORE~\cite{zhao2020smore} and SSGNN~\cite{sui2022scan} avoid external training data but similarly lack mechanisms for cross-view correspondence.

We therefore propose a reference- and memory-based method, \textbf{STRMSR}, for the SR step of cardiac MRI SVR. Given a target LR volume as a video, we leverage auxiliary HR reference views from the same subject to guide single-frame through-plane SR and take the intermediate SR outputs as the memory to propagate SR results along the third axis. Our contributions are summarized as follows:
(1) We propose coarse-to-fine correspondence matching and patch-wise dynamic feature aggregation to enable robust detail transfer between target and reference/memory images under cross-view geometric misalignment.
(2) We introduce a memory-based inter-slice SR propagation mechanism inspired by recent advances in video object segmentation~\cite{oh2019video,cheng2021rethinking,ye2025continuous}, which enforces slice-to-slice consistency and improves volumetric coherence in the reconstructed 3D HR output.
(3) We validate the proposed method on the WHS cardiac MRI dataset under two reference protocols to demonstrate its effectiveness across distinct geometries and reference coverage densities.

\section{Method}

\begin{figure*}[t]
    \centering
    \includegraphics[width=1.0\textwidth]{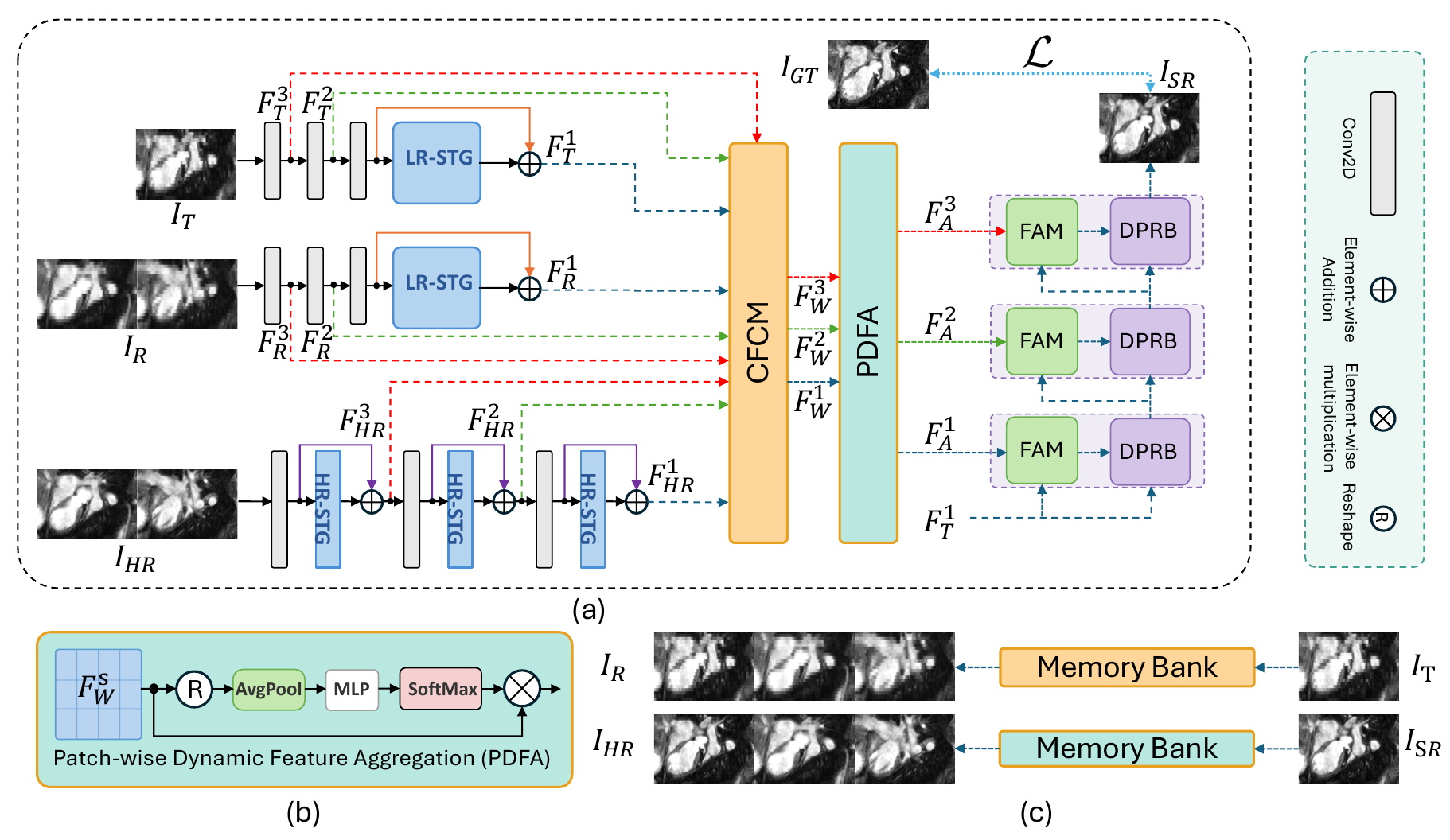}
    \caption{(a) Overall architecture of STRMSR, where we perform coarse-to-fine contextual matching (CFCM) and (b) patch-wise dynamic feature aggregation (PDFA). (c) The intermediate super-resolution results are stored in the memory bank. %\textcolor{blue}{Combine $L_{dc}$ and $L_{dc}$ as $L$ and use $\mathcal{L}$. In (c), please add in $I_T$ and $I_{SR}$ on the right most. Also, (b) should align with (a) on the left side.}
    }
    %Three parallel branches extract multi-scale features from the target LR ($I_{T}$), reference LR ($I_{R}$), and reference HR images ($I_{HR}$). CFCM performs coarse-to-fine contextual correspondence learning, and PDFA aggregates warped features from multiple references and memory frames. Progressive HR image reconstruction is performed via Feature Alignment Module (FAM) and Dual-Path Refinement Block (DPRB).}
    \label{fig:architecture}
\end{figure*}

%In local image regions, neighboring pixels tend to originate from the same object and exhibit strong similarity, which enables reliable local correspondence learning and highlights accurate feature representation as the key to reference-based SR. 
%Moreover, different reference views provide complementary anatomical information that is crucial for high-quality HR image reconstruction. 
%For reference-based SR, resolution and spatial misalignment between the target LR and reference HR images makes direct feature transfer unreliable. 
%We first downsample the reference (or memory) HR image ($I_{HR}$) into the same through-plane resolution as the target LR image ($I_{T}$), denoting as $I_{R}$. We then propose a hierarchical matching scheme that progressively refines feature alignment from coarse to fine levels and a patch-wise dynamic feature aggregation module for dynamic information fusion. Below, we give details of our method.
%, avoiding accumulated errors from simple interpolation-based feature warping~\cite{li2022transformer}.
%As shown in Fig.~\ref{fig:architecture}, the proposed STRMSR network improves through-plane resolution of the target LR volume using cross-view HR guidance. We use three parallel branches for feature extraction (target LR, reference LR, reference HR). Multi-scale features are then fed into CFCM to obtain matched reference features across scales and views, followed by PDFA to aggregate features. Finally, multi-scale feature aggregation blocks guide HR reconstruction to obtain the target SR images.

\subsection{Dual-Branch Transformer for Feature Extraction}
We first downsample the reference (or memory) HR image ($I_{HR}$) into the same through-plane resolution as the target LR image ($I_{T}$), denoting as $I_{R}$.
As shown in Fig.~\ref{fig:architecture}(a), we then extract target and reference features ($F_{T}$ and $F_{R}$) using a dual-branch encoder based on Swin Transformer group (STG), consisting of Residual Swin Transformer Blocks (RSTB). Target LR and reference LR branches share parameters to keep their embeddings in a common feature space, while the reference HR images are encoded by an independent branch to preserve resolution-specific details.
For LR inputs ($I_{T}$ and $I_{R}$), we first apply center-copy upsampling along the through-plane axis to match the HR reference resolution, then extract features $F_{T/R}^{s}$ at finer levels ($s=2,3$) using strided convolutions ($stride=2$). Lastly, we apply RSTB only at the coarsest level ($s=1$):
\begin{equation}
    F_{T/R}^{1} = \mathrm{RSTB}_{\mathrm{LR}}(F_{T/R}^{2}) + F_{T/R}^{2}.
\end{equation}
For HR reference ($I_{HR}$), we encode features $F_{HR}$ at all levels using RSTB blocks:
\begin{equation}
    F_{HR}^{s} = \mathrm{RSTB}_{\mathrm{HR}}^{s}(F_{HR}^{s+1}) + F_{HR}^{s+1}, \quad s\in\{1,2,3\}, \quad F_{HR}^{4}=I_{HR}.
\end{equation}
%where $F_{HR}^{0}$=$I_{HR}$.
%These multi-scale features $\{F_T^s,F_R^s,F_{HR}^s\}_{s\in\{1,2,3\}}$ are then used for contextual matching as detailed in Sec.~\ref{sec:ctf-matching}.

\subsection{Coarse-to-Fine Contextual Matching (CFCM)}
\label{sec:ctf-matching}

\begin{figure}[t]
    \centering
    \includegraphics[width=1.0\textwidth]{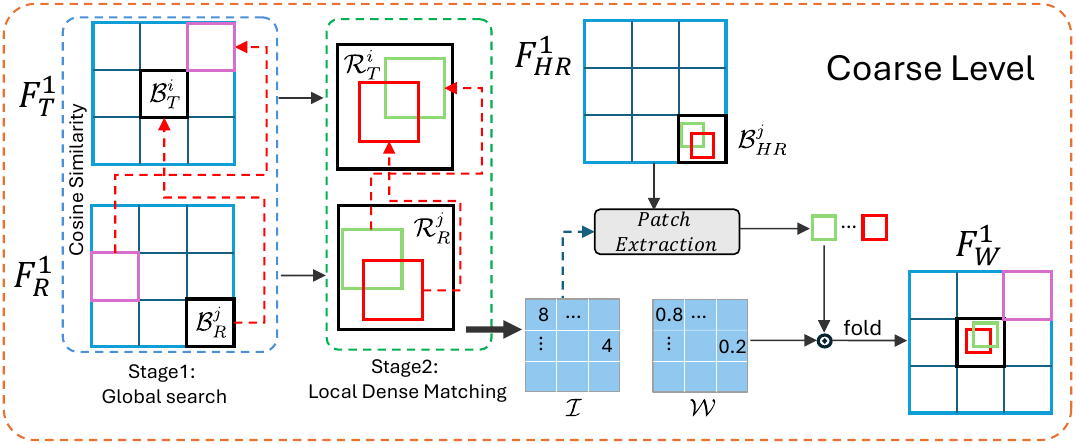}
    \caption{Illustration of CFCM at the coarsest level
    ($s{=}1$). %\textcolor{blue}{Please use $\mathcal{I}$, $\mathcal{W}$.}
    }
    %, target blocks $B_T^k$ are matched to reference blocks
    %$B_R^k$ via multi-dilation correlation (Stage~1), followed by dense
    %$3{\times}3$ patch matching within a local crop (Stage~2) to obtain
    %index map $\mathcal{I}_i$ and confidence map $\mathcal{W}_i$; HR patches
    %are assembled via confidence-weighted folding to produce $F_W^1$.}
    %\textbf{Bottom:} At finer levels, matched centers propagate via
    %$\mathrm{NN}$ interpolation and local dense matching refines
    %correspondence to yield $F_W^{s+1}$ ($s{+}1\in\{2,3\}$).}
    \label{fig:cfcm}
\end{figure}

%In local image regions, neighboring pixels tend to originate from the same object and exhibit strong similarity, which enables reliable local correspondence learning and highlights accurate feature representation as the key to reference-based SR. Moreover, different reference views provide complementary anatomical information that is crucial for high-quality HR image reconstruction. However, spatial misalignment between the target LR and reference HR images makes direct feature transfer unreliable. The motivates us to propose a hierarchical matching scheme that progressively refines alignment from coarse to fine levels, avoiding accumulated errors from simple interpolation-based feature warping~\cite{li2022transformer}.

Given multi-scale feature pyramids $\{ F_T^s, F_R^s, F_{HR}^s\}_{s \in \{1, 2, 3\}}$ at resolutions $H/4$, $H/2$, and $H$, 
%extracted from the target LR, reference LR, and reference HR images respectively, 
we aim to build dense correspondence between $F_T^s$ and $F_R^s$, then warp the HR features $F_{HR}^s$ accordingly. To ensure spatial consistency during center propagation across levels, a strict pyramid is enforced by padding the coarsest level when required and defining finer levels as exact $2\times$ and $4\times$ upscalings.
%As illustrated in Fig.~\ref{fig:cfcm}, CFCM establishes this via global coarse matching at $s{=}1$ followed by local hierarchical refinement at $s\in\{2,3\}$.

As illustrated in Fig.~\ref{fig:cfcm}, at the coarsest level ($s=1$), we partition $F_T^1$ into $M$ non-overlapping $k \times k$ blocks ($\mathcal{B}$). For each target block $i$ in $F_{T}^1$, we compute multi-dilation normalized correlation~\cite{lu2021masa} with all blocks in the reference feature $F_{R}^1$ to locate its best match at block $j$. We then crop a local search region $\mathcal{R}$ of size $d \times d$ centered at $i$ and $j$.
Within each local search region pair, we perform dense $3 \times 3$ patch matching. For each target patch $p_u$ and reference patch $q_v$, we compute cosine similarity~\cite{li2022transformer,lu2021masa} and obtain correspondence:
 \begin{equation}
     \mathcal{S}_{uv} = \langle p_u / \|p_u\|, \, q_v / \|q_v\| \rangle, \quad
     \mathcal{I}_u = \arg\max_v \mathcal{S}_{uv}, \quad 
     \mathcal{W}_u = \max_v \mathcal{S}_{uv},
 \end{equation}
 where $\mathcal{I}$ is the index map indicating the best matching position, and $\mathcal{W}$ is the confidence map. 
 %With index maps $\mathcal{I}$, 
 We extract corresponding $3 \times 3$ patches from the reference HR feature $F_{HR}^s$ via $\mathcal{I}$ and assemble them via confidence-weighted folding:
\begin{equation}
    F_W^1 = \mathrm{fold}\left(\{\mathcal{W}_u \cdot F_{HR}^1[\mathcal{I}_u] \}_{i}\right),
\end{equation}
where $[\cdot]$ denotes patch extraction, $\mathrm{fold}$ performs the inverse of unfolding with overlap averaging, and $F_W^1$ denotes the warped HR reference feature.

Different from McMRSR~\cite{li2022transformer}, which mainly reuses the LR-level correspondence for multi-scale feature transfer, our CFCM uses it only as an initialization and further refines the correspondence at finer levels $s \in {2, 3}$. The matched block centers $\mathcal{C}_{i/j}$ propagate from level $s$ to $s+1$ via:
\begin{equation}
\mathcal{C}_{i/j}^{s+1} = 2 \cdot \mathrm{NN}(\mathcal{C}_{i/j}^{s}),
\end{equation}
where nearest-neighbor interpolation ($\mathrm{NN}$) preserves block boundaries. At each finer level, we perform a new dense local search within radius $r$ around the propagated center to update the correspondence, producing warped HR reference features $F_W^2$ and $F_W^3$.

\subsection{Patch-wise Dynamic Feature Aggregation (PDFA)}
Given $Z$ warped feature maps $\{F_{W,z}^s\}_{z=1}^{Z}$ at scale $s\in\{1,2,3\}$ obtained from  $Z$ multiple HR references and propagated memory frames, the goal is to fuse them into a single spatially aligned feature $F_A^s$. Direct pixel-wise gating is expensive and tends to overfit to local misalignments.
We therefore propose a patch-wise dynamic aggregation strategy, as shown in Fig.~\ref{fig:architecture}(b). 
%The key assumption is that the reliability of correspondence varies smoothly within a local neighborhood, i.e., boundaries and textured regions benefit from high-confidence reference cues, whereas homogeneous regions should be fused more conservatively. Accordingly, we predict one mixture weight per $p\times p$ patch and share it across all pixels in the patch, which reduces complexity and stabilizes the fusion.
Formally, we split each $F_{W,z}^s \in \mathbb{R}^{C\times H\times W}$ into non-overlapping $p\times p$ patches indexed by $h$, and compute a patch descriptor via average pooling:
\begin{equation}
    d_{z,h} = \mathrm{AvgPool}(F_{W,z,h}^s) \in \mathbb{R}^{C}.
\end{equation}
A lightweight MLP $g(\cdot)$ then produces a patch logit $\ell_{z,h}=g(d_{z,h})$. Finally, we use softmax to normalize logits across the $Z$ candidates and broadcast weights to pixel resolution within the patch, yielding the fused feature:
\begin{equation}
    \alpha_{z,h} = \frac{\exp(\ell_{z,h})}{\sum_{j=1}^{Z} \exp(\ell_{j,h})}, \qquad
    F_A^s(x,y) = \sum_{z=1}^{Z} \alpha_{z,h(x,y)} \, F_{W,z}^s(x,y),
\end{equation}
where $h(x,y)$ denotes the patch index containing pixel $(x,y)$. This formulation can be interpreted as a content-adaptive mixture-of-experts~\cite{chen2025heterogeneous,zamfir2024see} over warped references/memories, enabling selective transfer of informative patches while suppressing inconsistent matches.

\subsection{Progressive Multi-Scale Feature Fusion and Reconstruction}
Given fused HR reference/memory features $\{F_A^s\}_{s \in \{1,2,3\}}$ aggregated from multiple reference/memory frames, we reconstruct the target HR image by combining feature alignment and detail refinement via Feature Alignment Module (FAM) and Dual-Path Refinement Block (DPRB) as in~\cite{li2022transformer}.

\subsection{Memory-based SR Propagation}
We take the LR volume as a video, where the third axis is deemed as the temporal axis. As shown in Fig.~\ref{fig:architecture}(c), once we use the reference images to super resolve the first frame, we store the $I_{T}$-$I_{SR}$ pair in a First-In-First-Out (FIFO) memory bank. The reference images together with memory images can then guide SR of following frames, where the results can be further stored in the memory bank to improve slice-to-slice SR consistency. 

\subsection{Loss Function}
We optimize the network using a combination of an image-domain $\ell_1$ reconstruction loss and a $k$-space loss:
$\mathcal{L} = \lambda_{\mathrm{rec}}\left\| I_{SR} - I_{GT} \right\|_{1} + \lambda_{\mathrm{k}}\left\| \mathcal{F}(I_{SR}) - \mathcal{F}(I_{GT}) \right\|_{2}^{2}$,
%\begin{equation}
    %\mathcal{L} = \lambda_{\mathrm{rec}}\underbrace{\left\| I_{SR} - I_{GT} \right\|_{1}}_{\mathcal{L}_{\mathrm{rec}}} + \lambda_{\mathrm{dc}}\underbrace{\left\| \mathcal{F}(I_{SR}) - \mathcal{F}(I_{GT}) \right\|_{2}^{2}}_{\mathcal{L}_{\mathrm{dc}}},
%\end{equation}
%\begin{equation}
    %\mathcal{L} = \lambda_{\mathrm{rec}}\left\| I_{SR} - I_{GT} \right\|_{1} + \lambda_{\mathrm{dc}}\left\| \mathcal{F}(I_{SR}) - \mathcal{F}(I_{GT}) \right\|_{2}^{2},
%\end{equation}
where $I_{GT}$ is the ground truth HR image, $\mathcal{F}(\cdot)$ denotes the 2D FFT applied slice-wise, and we set $\lambda_{\mathrm{rec}}=1$ and $\lambda_{\mathrm{k}}=0.001$ in all experiments.

\section{Experiments}

\subsection{Datasets and Data Preprocessing}

We evaluate STRMSR on the WHS cardiac MRI dataset~\cite{li2022atrialjsqnet,gao2023bayeseg} under two SR protocols. 
We split the 86 subjects into 50, 6, 30 as the training, validation and testing set, respectively.
(1)~\textbf{\textit{Orthogonal-plane reference protocol (WHS-Ortho).}} To assess STRMSR under dense reference coverage, we adopted the multi-plane acquisition scheme of~\cite{kuklisova2012reconstruction} on the same WHS volumes (resampled to $256\times256\times176$). The axial orientation was treated as the target. Images were degraded through-plane by Gaussian pre-filtering ($\sigma=1.0$) and $\times4$ or $\times8$ block averaging, yielding LR volumes.
%of $256\times256\times22$ at $\times8$. 
The sagittal orientation was degraded along its own through-plane axis to produce an auxiliary reference volume that retained HR in-plane resolution along the target's through-plane direction. 
%Compared with the long-axis protocol, this orthogonal-plane configuration provides substantially denser reference coverage of the target stack. 
This orthogonal-plane configuration is the same as slice-to-volume reconstruction used in fetal MRI~\cite{kuklisova2012reconstruction} except that we only used one reference volume.
The two nearest neighboring slices from the sagittal reference volume were used as references for each target slice.
(2)~\textbf{\textit{Long-axis reference protocol (WHS-LAX).}} Following~\cite{xu2024improved}, segmentation masks were used to extract anatomical landmarks and construct a standardized short-axis (SAX) cardiac coordinate system. Volumes were resampled to $256^3$ voxels, and long-axis (LAX) 2-, 3-, 4-chamber (CH) views were generated as the HR references. Degradation was performed through Gaussian pre-filtering ($\sigma=1.0$) followed by $\times4$ or $\times8$ through-plane downsampling via block averaging to generate the SAX LR images as well as the LAX HR reference images. 
This long-axis configuration is the same as clinical cardiac cine MRI except that we included in the atrium area for the SAX view and we only covered a single cardiac phase.
Starting with the central slice (4-CH view), we performed through-plane SR along the third axis bidirectionally towards the two ends.
For both protocols, intensities were normalized using $1st$-$99th$ percentiles. 

\subsection{Implementation Details and Baseline Methods}
STRMSR was implemented in PyTorch and trained on NVIDIA A100 GPUs (80\,GB) using the Adam optimizer with a learning rate of $1 \times 10^{-4}$ and a batch size of 2 for 250 epochs. The network employs 4 RSTB blocks.
%(2 Swin-Transformer Layer each) with window size 8, embedding dimension 60, and MLP ratio 2.67. 
The block size is $k{=}8$, local search region $d{=}13$, 
and PDFA patch size $p{=}8$. Local refinement radii are $r{=}4$ 
and $r{=}5$ at levels $s{=}2$ and $s{=}3$, respectively. %For the memory bank, during training, we store $T{=}2$ most recent SR results for WHS-LAX and $T{=}3$ for WHS-Ortho; during inference, the bank size is expanded to $T{=}10$ for WHS-LAX and $T{=}3$ for WHS-Ortho.
For the memory bank, during training, we store $T{=}3$ most recent SR results for WHS-Ortho and $T{=}2$ for WHS-LAX; 
during inference, the bank size is expanded to $T{=}3$ for WHS-Ortho and $T{=}10$ for WHS-LAX.
%Separate LR and HR RSTB branches with unshared weights are used. The training loss combines $\ell_1$ reconstruction and $k$-space data consistency losses ($\lambda_{\mathrm{rec}}=1$, $\lambda_{\mathrm{dc}}=10^{-4}$). 
Performance was evaluated using PSNR (dB), SSIM and MSE under $\times 4$ and $\times 8$ through-plane SR on both protocols.

\subsection{Baseline Methods}
Bicubic is a single-image interpolation baseline. MsFF-Net~\cite{kang20243d} is a 3D multi-resolution CNN that fuses an auxiliary HR reference volume with the LR target. MINet~\cite{feng2021multi} takes a single paired reference: the nearest sagittal slice for WHS-Ortho and the 4-CH view for WHS-LAX. MASA~\cite{lu2021masa} and McMRSR~\cite{li2022transformer} support multi-reference fusion and use the same references as our method, i.e., the two nearest sagittal HR slices for WHS-Ortho and the 2-, 3-, 4-CH LAX HR views for WHS-LAX. All baselines were retrained on our datasets under identical settings.

\subsection{Results}

\begin{table}
\caption{Quantitative comparison for through-plane SR on WHS under WHS-Ortho and WHS-LAX protocols. 
Values are mean(std); \textcolor{red}{red} and \textcolor{blue}{blue} mark the best and second-best. Symbols on STRMSR indicate Wilcoxon signed-rank significance over all baselines ($^{\dagger}p<0.001$).
}\label{tab:comparison}
\centering
\setlength{\tabcolsep}{3pt}
\resizebox{\textwidth}{!}{%
\begin{tabular}{c|c|c|c|c|c|c|c}
\hline
\multirow{2}{*}{Scale} & \multirow{2}{*}{Method} & \multicolumn{3}{c|}{WHS-Ortho~\cite{li2022atrialjsqnet,gao2023bayeseg}} & \multicolumn{3}{c}{WHS-LAX~\cite{li2022atrialjsqnet,gao2023bayeseg}} \\
\cline{3-8}
& & PSNR & SSIM & MSE ($\times 10^{-3}$) & PSNR & SSIM & MSE ($\times 10^{-3}$) \\
\hline
$\times$4 & Bicubic & 29.96(2.80) & 0.9175(0.0332) & 1.228(0.838) & 27.14(1.48) & 0.8450(0.0360) & 2.046(0.696) \\
$\times$4 & MsFF-Net (3D)~\cite{kang20243d} & \textcolor{blue}{37.38(3.55)} & \textcolor{red}{0.9747(0.0147)} & 0.245(0.178) & \textcolor{blue}{29.83(1.59)} & \textcolor{blue}{0.8891(0.0332)} & \textcolor{blue}{1.115(0.443)} \\
$\times$4 & MINet~\cite{feng2021multi} & 35.83(3.15) & 0.9643(0.0195) & 0.325(0.205) & 29.49(1.54) & 0.8830(0.0331) & 1.196(0.453) \\
$\times$4 & MASA~\cite{lu2021masa} & 37.13(3.01) & 0.9706(0.0172) & \textcolor{blue}{0.239(0.155)} & 29.77(1.60) & 0.8830(0.0332) & 1.128(0.450) \\
$\times$4 & McMRSR~\cite{li2022transformer} & 37.17(3.25) & 0.9676(0.0179) & 0.243(0.166) & 29.81(1.61) & 0.8852(0.0336) & 1.119(0.451) \\
$\times$4 & STRMSR (Ours) & \textcolor{red}{37.67(3.16)} & \textcolor{blue}{0.9721(0.0169)} & \textcolor{red}{0.219(0.155)} & \textcolor{red}{29.98(1.63)} & \textcolor{red}{0.8892(0.0331)} & \textcolor{red}{1.081(0.445)} \\
\hline
$\times$8 & Bicubic & 25.08(2.27) & 0.7852(0.0586) & 3.544(1.959) & 23.74(1.43) & 0.7198(0.0491) & 4.455(1.476) \\
$\times$8 & MsFF-Net (3D)~\cite{kang20243d} & \textcolor{blue}{29.09(1.79)} & \textcolor{blue}{0.8722(0.0375)} & \textcolor{blue}{1.329(0.477)} & 25.45(1.24) & 0.7629(0.0489) & 2.962(0.843) \\
$\times$8 & MINet~\cite{feng2021multi} & 27.99(1.79) & 0.8488(0.0435) & 1.703(0.585) & 25.45(1.23) & 0.7659(0.0481) & 2.963(0.841) \\
$\times$8 & MASA~\cite{lu2021masa} & 28.48(1.71) & 0.8550(0.0427) & 1.512(0.501) & 25.48(1.25) & 0.7626(0.0482) & 2.947(0.844) \\
$\times$8 & McMRSR~\cite{li2022transformer} & 28.67(1.76) & 0.8537(0.0432) & 1.453(0.496) & \textcolor{blue}{25.68(1.28)} & \textcolor{blue}{0.7674(0.0498)} & \textcolor{blue}{2.819(0.823)} \\
$\times$8 & STRMSR (Ours) & \textcolor{red}{29.64(1.80)}\textsuperscript{$\dagger$} & \textcolor{red}{0.8764(0.0406)}\textsuperscript{$\dagger$} & \textcolor{red}{1.174(0.439)}\textsuperscript{$\dagger$} & \textcolor{red}{25.87(1.31)}\textsuperscript{$\dagger$} & \textcolor{red}{0.7773(0.0499)}\textsuperscript{$\dagger$} & \textcolor{red}{2.702(0.814)}\textsuperscript{$\dagger$} \\
\hline
\end{tabular}%
}
\end{table}

\begin{figure}
    \centering
    \includegraphics[width=1.0\linewidth]{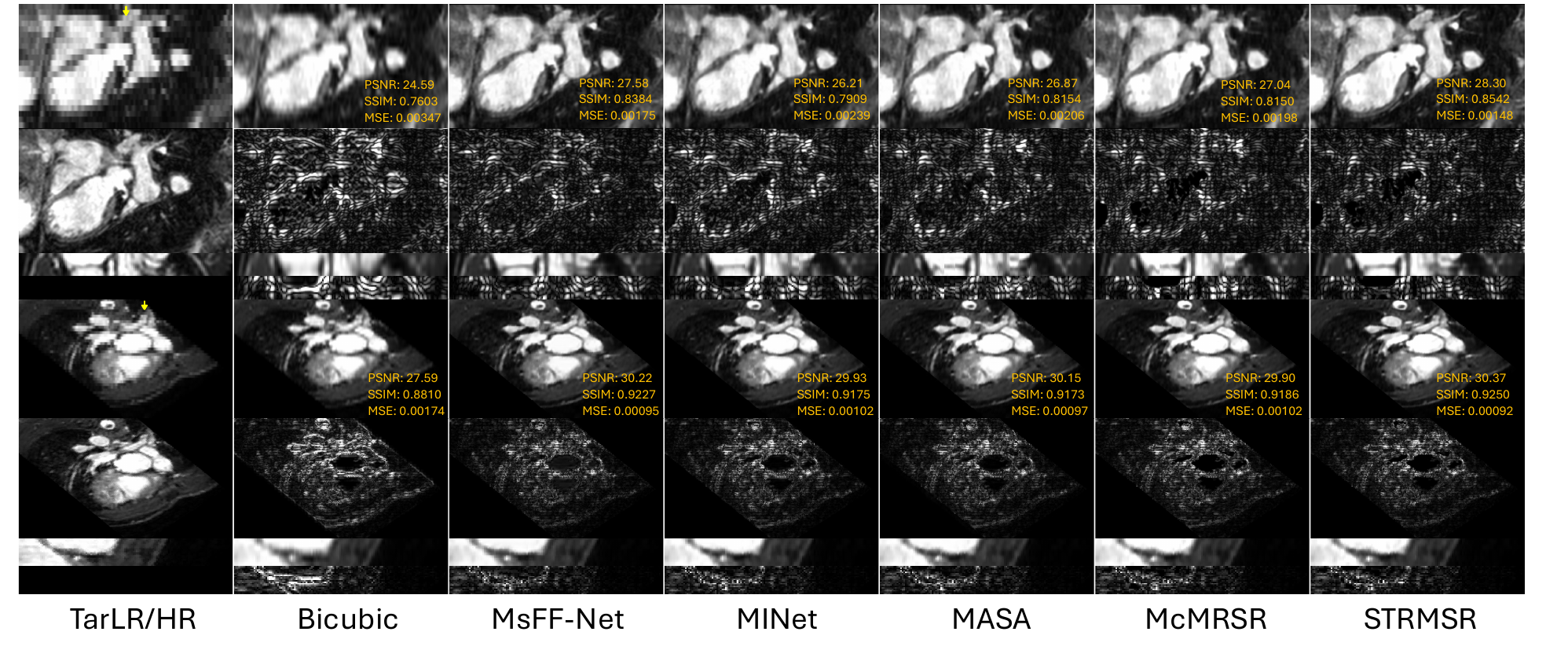}
    \caption{Qualitative comparison under $\times$8 SR on WHS-Ortho (rows 1--4) and $\times$4 SR on WHS-LAX (row 5--8). Rows 1, 5: SR images; rows 2, 6: error maps; rows 3, 7: temporal profiles at yellow arrows; rows 4, 8: temporal error maps.
    }
    \label{fig:qualitative}
\end{figure}

\textbf{\textit{Quantitative Comparison.}}
As shown in Table~\ref{tab:comparison}, STRMSR achieves the best performance on both protocols at $\times 8$, and the best on five out of six metrics at $\times 4$. Bicubic lacks external cues and degrades sharply from $\times 4$ to $\times 8$. MsFF-Net, despite leveraging both 3D context and an HR reference, processes the reference only implicitly through multi-resolution fusion without explicit cross-view correspondence, limiting its detail transfer; it slightly surpasses STRMSR on $\times 4$ WHS-Ortho SSIM but trails clearly at $\times 8$ on both protocols. MINet fuses references implicitly with a single paired view, while MASA and McMRSR employ explicit matching but operate on single frames without inter-slice propagation, leaving volumetric consistency unaddressed (see temporal profiles in Fig.~\ref{fig:qualitative}). STRMSR addresses both via coarse-to-fine matching and memory-based propagation. The performance gap is larger on WHS-Ortho than WHS-LAX because the sagittal reference densely covers every axial slice, whereas the few LAX views provide far less reference information per target slice. The gap also grows from $\times 4$ to $\times 8$: at $\times 4$, abundant reference cues let MsFF-Net match STRMSR closely (no significant difference on any metric by Wilcoxon signed-rank test), while STRMSR still significantly outperforms all other baselines including MASA and McMRSR ($p<0.001$). At $\times 8$, the reference becomes sparse and only models that exploit limited information remain effective---STRMSR improves PSNR by $+0.97$\,dB over McMRSR and $+0.55$\,dB over MsFF-Net on WHS-Ortho, with all $\times 8$ improvements significant at $p<0.001$. This is a practically relevant regime, as $\times 8$ through-plane spacing is common in clinical cardiac MRI to meet breath-hold constraints, demonstrating that STRMSR is the most useful where the problem is the hardest.

\textbf{\textit{Qualitative Comparison.}}
Fig.~\ref{fig:qualitative} presents visual results on WHS-Ortho ($\times$8) and WHS-LAX ($\times$4). On WHS-Ortho, Bicubic produces severe blurring, and MsFF-Net, despite leveraging 3D context and an HR reference, still over-smooths boundaries due to its implicit multi-resolution fusion, similar to MINet. MASA and McMRSR recover partial structure via reference matching, but their error maps show significant residual blurring along myocardial walls due to the lack of inter-slice propagation. STRMSR produces the sharpest edges and lowest error, and its temporal profile clearly preserves a thin linear structure that all competing methods either blur or miss, confirming the benefit of memory-based propagation. On WHS-LAX, the gap narrows under milder $\times$4 degradation, and although the long-axis views provide substantially less reference information per target slice than the dense sagittal volume in WHS-Ortho, STRMSR still achieves the best PSNR/SSIM/MSE and visibly cleaner myocardial boundaries, demonstrating its robustness under sparse reference coverage. %\textcolor{blue}{I don't think the temporal profile looks better for STRMSR under $\times$4 SR. You'd better select annother example to support your claim.}

\begin{table}[t]
\caption{Ablation study on WHS-Ortho ($\times$8 through-plane SR).}
\label{tab:ablation-whs-ortho-x8}
\centering
\setlength{\tabcolsep}{4pt}
\resizebox{\textwidth}{!}{%
\begin{tabular}{l|ccc|ccc}
\hline
\multirow{2}{*}{Variant} & \multicolumn{3}{c|}{Modules} & \multirow{2}{*}{PSNR} & \multirow{2}{*}{SSIM} & \multirow{2}{*}{MSE ($\times 10^{-3}$)} \\
\cline{2-4}
 & CFCM & PDFA & Memory &  &  & \\
\hline
w/o CFCM & $\times$ & $\checkmark$ & $\checkmark$ & 29.04(1.69) & 0.8643(0.0436) & 1.335(0.474) \\
w/o PDFA & $\checkmark$ & $\times$ & $\checkmark$ & 29.40(1.76) & 0.8708(0.0422) & 1.236(0.461) \\
w/o Memory & $\checkmark$ & $\checkmark$ & $\times$ & 29.52(1.81) & 0.8737(0.0425) & 1.209(0.468) \\
\textbf{Full (STRMSR)} & $\checkmark$ & $\checkmark$ & $\checkmark$ & \textbf{29.64(1.80)} & \textbf{0.8764(0.0406)} & \textbf{1.174(0.439)} \\
\hline
\end{tabular}%
}
\end{table}

\textbf{\textit{Ablation Study.}}
Table~\ref{tab:ablation-whs-ortho-x8} evaluates each component on WHS-Ortho under $\times$8 SR. Replacing CFCM with the McMRSR-style matching strategy causes the largest drop ($-0.60$,dB PSNR), confirming the benefit of our coarse-to-fine correspondence refinement over directly reused LR-level correspondences. Disabling PDFA reduces PSNR by $0.24$,dB, showing that content-adaptive aggregation helps suppress unreliable HR feature transfers. Removing the memory bank leads to a $0.12$,dB drop, indicating that inter-slice propagation provides a consistent gain for volumetric SR. Overall, the three components contribute complementarily, yielding the best performance when combined.

\section{Conclusion}
We presented STRMSR, a reference- and memory-guided framework for through-plane cardiac MRI SR. %By combining coarse-to-fine contextual matching with patch-wise dynamic feature aggregation, 
With CFCM and PDFA, STRMSR establishes robust cross-view correspondence under spatial misalignment and selectively transfers informative reference/memory details, while memory-based inter-slice propagation further enforces volumetric coherence. The framework generalizes across anatomical geometries with distinct reference coverage densities, suggesting broader applicability to other anisotropic MRI SR tasks.

\begin{comment}
\section{Conclusion}
We presented STRMSR for through-plane cardiac MRI SR. 
STRMSR can establish robust cross-view correspondence, selectively transfer informative reference/memory details, and enforce volumetric coherence. 
The framework generalizes across anatomical geometries with distinct reference coverage densities,
suggesting broader applicability to other anisotropic MRI SR tasks.

\begin{credits}
\subsubsection{\ackname} A bold run-in heading in small font size at the end of the paper is
used for general acknowledgments, for example: This study was funded
by X (grant number Y).
\subsubsection{\discintname}
The authors have no competing interests to declare that are relevant to the contents of this article.
\end{credits}
\end{comment}
c

\bibliographystyle{splncs04}
\bibliography{main}

% \end{thebibliography}
\end{document}